\documentclass{article}

\usepackage{arxiv}

\usepackage{amsmath,amssymb}
\usepackage{mathtools}
\usepackage{xspace}
\usepackage{algorithm}
\usepackage[noend]{algorithmic}
\usepackage{color}
\usepackage{hyperref}
\usepackage{graphicx}
\usepackage{tikz}
\usepackage{wasysym}
\usepackage{booktabs}
\usepackage{multirow}
\usepackage{float}
\usepackage{pdflscape}
\usepackage[normalem]{ulem}
\usepackage{makecell}
\usepackage{xcolor}
\usepackage{csquotes}
\usepackage{pifont}
\usepackage{colortbl}
\usepackage[export]{adjustbox}
\usepackage[absolute]{textpos}

\begin{document}

\title{Deep Learning as a Competitive Feature-Free Approach for Automated Algorithm Selection\\ on the Traveling Salesperson Problem}
%
%
\author{
  Moritz Seiler \\
  Statistics and Optimization\\
  University of M\"unster\\
  M\"unster, Germany \\
  \texttt{moritz.seiler@uni-muenster.de} \\
  \And
  Janina Pohl \\
  Statistics and Optimization\\
  University of M\"unster\\
  M\"unster, Germany \\
  \texttt{janina.pohl@uni-muenster.de} \\
  \And
  Jakob Bossek \\
  Optimisation and Logistics\\
  The University of Adelaide\\
  Adelaide, Australia \\
  \texttt{jakob.bossek@adelaide.edu.au} \\
  \And
  Pascal Kerschke \\
  Statistics and Optimization\\
  University of M\"unster\\
  M\"unster, Germany \\
  \texttt{kerschke@uni-muenster.de} \\
  \And
  Heike Trautmann \\
  Statistics and Optimization\\
  University of M\"unster\\
  M\"unster, Germany \\
  \texttt{trautmann@uni-muenster.de} \\
}

\maketitle

\begin{abstract}
In this work we focus on the well-known Euclidean Traveling Salesperson Problem (TSP) and two highly competitive inexact heuristic TSP solvers, EAX and LKH, in the context of per-instance algorithm selection (AS). We evolve instances with $1\,000$ nodes where the solvers show strongly different performance profiles. These instances serve as a basis for an exploratory study on the identification of well-discriminating problem characteristics (features). Our results in a nutshell: we show that even though (1) promising features exist, (2) these are in line with previous results from the literature, and (3) models trained with these features are more accurate than models adopting sophisticated feature selection methods, the advantage is not close to the virtual best solver in terms of penalized average runtime and so is the performance gain over the single best solver. However, we show that a feature-free deep neural network based approach solely based on visual representation of the instances already matches classical AS model results and thus shows huge potential for future studies.
\end{abstract}


\section{Introduction}
\label{sec:introduction}

The \emph{Traveling Salesperson Problem} (TSP) is a classical $\mathcal{NP}$-hard optimization problem of utmost relevance, e.g., in transportation logistics, bioinformatics or circuit board fabrication. The goal is to route a salesperson through a set of cities such that each city is visited exactly once and the tour is of minimal length.
In the past decades tremendous progress has been made in the development of high-performing heuristic TSP solvers. The local search-based \emph{Lin-Kernigham Heuristic} (LKH)~\cite{helsgaun2000LKH} and the genetic algorithm \emph{Edge-Assembly-Crossover} (EAX)~\cite{Nagata2013EAX}, along with their respective restart versions introduced in Kotthoff et al.~\cite{kotthoff2015}, undeniably pose the state-of-the-art in inexact TSP solving.

\emph{Automated Algorithm Selection} (AS), originally proposed by Rice~\cite{Rice1976AS} back in 1976, is a powerful framework to predict the best-performing solver(s) from a portfolio of candidate solvers by means of machine learning. It has been successfully applied to a wide spectrum of challenging optimization problems in both the combinatorial~\cite{kotthoff2014,lindauer2017autofolio,malitsky2013_algorithm,rizzini2017,xu2012} 
and continuous domain~\cite{Kerschke2019ASonContinuous,bischl2012} with partly astonishing performance gains -- see the recent survey by Kerschke et al.~\cite{KerschkeHNT2019AutomatedAlgorithm} for a comprehensive overview. In particular, the TSP was subject to several successful AS-studies~\cite{kotthoff2015,KKBHTLeveragingTSP,MBB2012LocalSearchTSP,Mersmann2013,Pihera14} which exploited the complementary performance profiles of simple heuristics on the one hand and the state-of-the-art solvers LKH and EAX on classical TSP benchmark sets on the other hand.

In the classic setting, AS relies on characteristic problem instance properties, termed \emph{(instance) features}. These features are used as predictor variables for classical machine learning algorithms, e.g., random forests or support vector machines. The key idea -- and ideal outcome -- is that these features can easily be used to automatically derive decision rules that are well-suited to partition the instance space into ideally disjoint sub-spaces of instances, and which then are uniquely solved best by different solvers. However, features have many drawbacks: they are usually hand-crafted in a tedious process~\cite{Hutter2014AlgorithmRuntimePrediction}, partly require time-consuming calculations (which need to be taken into account by the model fitting step) and are problem-tailored (or at least specific to a problem domain, e.g., graph problems). Moreover, we usually prefer light-weight models with few features. Hence, training AS models is frequently combined with automated feature selection methods \cite{guyon2006introduction,peng2005feature,urbanowicz2018relief} or dimensionality reduction techniques \cite{Haer15}.

\begin{figure}[t]
    \centering
    \includegraphics[width=0.6\textwidth]{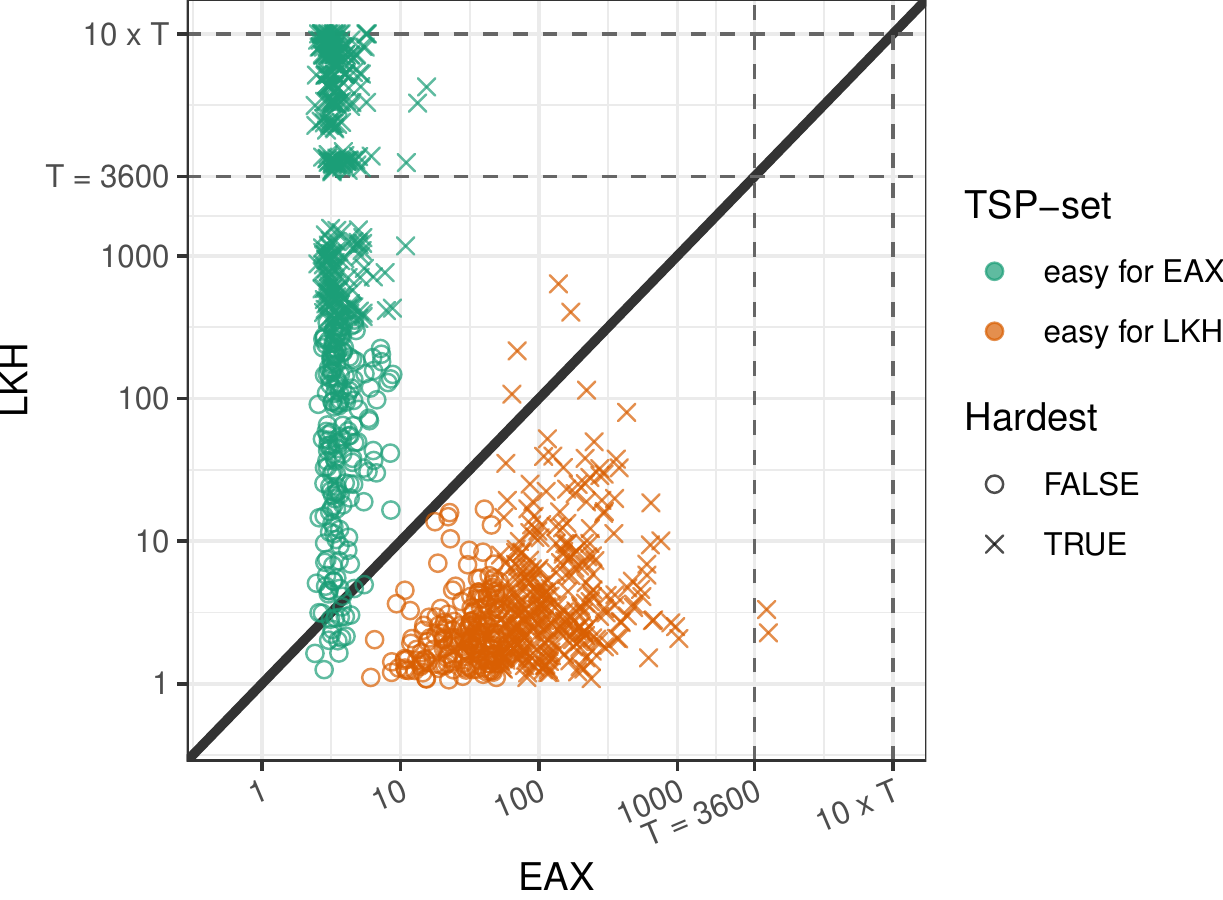}
    \caption{PAR10 values (log-scaled) of EAX and LKH show complementary performance on the two subsets of instances (easy for EAX or LKH) and thereby suggest huge potential for automated algorithm selection.}
    \label{fig:scatterplot_performance}
\end{figure}

Recently, Alissa et al.~\cite{Alissa2019ASWithoutFeatureExtraction} took a promising new path after exploring first approaches to avoid manual feature-crafting~\cite{Ross2004HyperHeuristics,Sim2012HyperHeuristicClassifier}. The authors proposed a deep learning based approach which does not rely on any a-priori calculated feature profiles. Instead, in their study on the 1D-Bin-Packing Problem (BPP) the neural network is given temporal sequence data of the BPP instance as the only input. They were able to achieve drastic improvements.
In this paper we adopt and adapt this idea for the TSP. To this end, we evolve a set of instances where LKH and EAX show strongly different behaviour in terms of \emph{Penalized Average Runtime}\footnote{The PAR10-score is a common measure in AS for combinatorial optimization problems. For a stochastic algorithm $A$ and an instances $I$ it is defined as the average of the running times of $A$ on $I$ where runs which did not reach the optimum within a given time limit $T$ are penalised by a factor of $10\cdot T$.} (PAR10; \cite{Bischl2016aslib}). We show that with classical AS we can clearly beat the \emph{Single Best Solver} (SBS; the solver with best average performance). However, the gap to the \emph{Virtual Best Solver} (VBS; perfectly predicting oracle) can only be reduced slightly with much room for improvement. This holds true even in the case when we enrich the machine learning pipeline with (1) hand-selected feature subsets (based on exploratory data analysis), (2) different feature selection methods, or (3) a combination of both. After that, we propose a feature-free deep learning approach where the neural networks are trained on the plain image representations of Euclidean TSP instances. This approach achieves competitive performance, but drops the need for manual feature derivation and calculation.

The remainder of this paper is structured as follows. We describe the benchmark set (generation) and pre-selection of feature subsets in Sections \ref{sec:evolving_instances} and \ref{sec:feature_identification} respectively. In Section~\ref{sec:as_classical} we present the results that we achieve using classical feature-based AS approaches. Next, in Section~\ref{sec:as_dnn}, we detail our feature-free deep learning based approaches, and compare the results with the classical models. We close with a discussion and an outlook in Section~\ref{sec:conclusion}.


\section{Evolving TSP Instances}
\label{sec:evolving_instances}

Our benchmark requires a set of Euclidean TSP instances that show strong differences in algorithmic performance. To this end, we adopt an \emph{evolutionary algorithm}~(EA) and creative mutation operators recently proposed by Bossek et al.~\cite{bossek2019Evolving}. Their method allows for the tailored generation\footnote{Various TSP benchmark libraries exist, e.g., TSPLIB~\cite{Reinelt91tsplib} or classical Random Uniform Euclidean~(RUE) instances. However, these instances are either inhomogeneous in size or exhibit very little structural difference. For the purpose of algorithm selection though a balanced and homogeneous benchmark set is highly beneficial.} of TSP instances that (1) have the desired performance difference (in terms of the ratio of PAR10-scores), (2) show multifarious topologies in terms of point arrangement in the Euclidean plane\footnote{The mutation operators are designed to evolve structures that can be observed in real-world TSP instances, e.g., \emph{Very Large Scale Integration} (VLSI) and are thus closer to the real-world than the often used random uniform problems.}, and (3) are well distributed in the space of instances characteristics/features; in particular the latter two properties were not achieved by evolutionary instance generation methods before~\cite{MBB2012LocalSearchTSP,Mersmann2013}. For sake of brevity we refer the reader to Bossek et al.~\cite{bossek2019Evolving} for more details.

We generated a balanced data set of $1\,000$ TSP instances with $n=1\,000$ nodes per instance using the EA parameters from \cite{bossek2019Evolving}; each $500$ being uniquely faster solved to optimality by either EAX or LKH.\footnote{We work with the restart versions of EAX and LKH which trigger a restart once the internal stopping conditions are met~\cite{kotthoff2015} as long as the time limit is not reached.}
All generated data is available in a public GitHub repository (\url{https://github.com/mvseiler/PPSN_TSP_DL}).

Fig.~\ref{fig:scatterplot_performance} depicts the performance -- measured by means of PAR10 -- of EAX and LKH on the entire benchmark set. The plot highlights apparent potential for automated algorithm selection due to strong performance differences. On the other hand the data reveals the general superiority of the EAX solver since it is much harder to evolve instances that are hard for EAX (shown as orange points). There are just two instances for which the EAX hits the cutoff time $T=3\,600$ seconds~(1 hour) at least once in all of its ten independent runs on that instance. In contrast, LKH frequently gets stuck in local optima -- see the cluster of green points between $T=3\,600$ and $10 \cdot T$ in the top left corner.

\section{Identifying Adequate Subsets of TSP Features}
\label{sec:feature_identification}
\begin{figure}[ht]
    \centering
    \includegraphics[width=0.32\textwidth,frame]{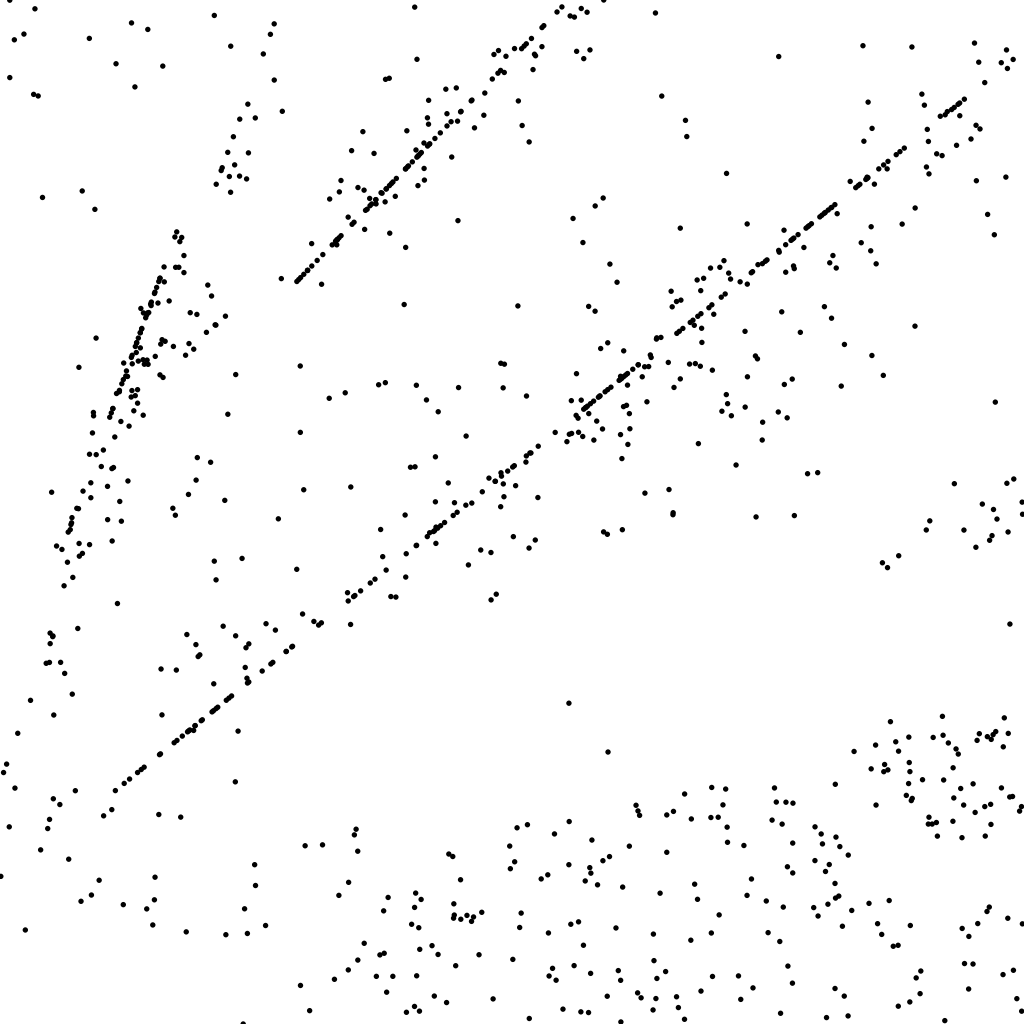}
    \hfill
    \includegraphics[width=0.32\textwidth,frame]{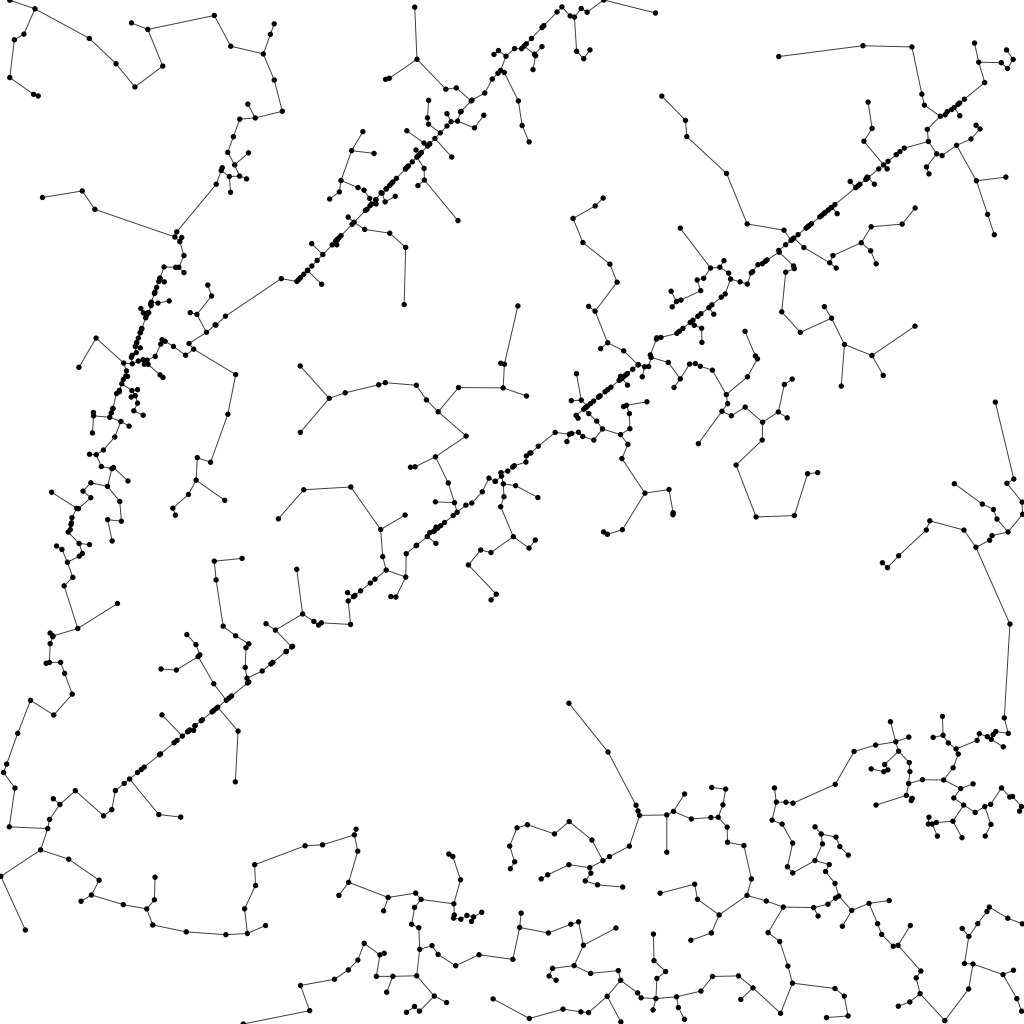}
    \hfill
    \includegraphics[width=0.32\textwidth,frame]{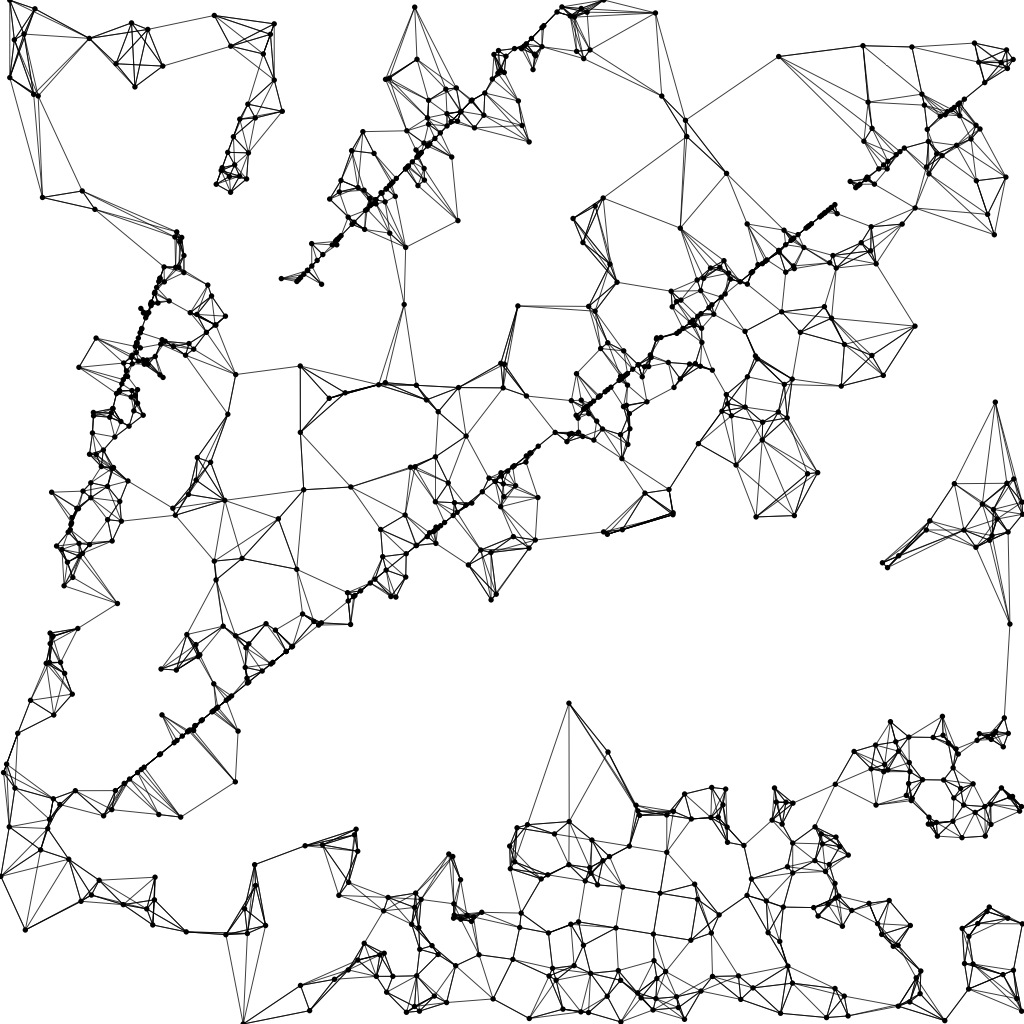}\\
    \includegraphics[width=0.32\textwidth,frame]{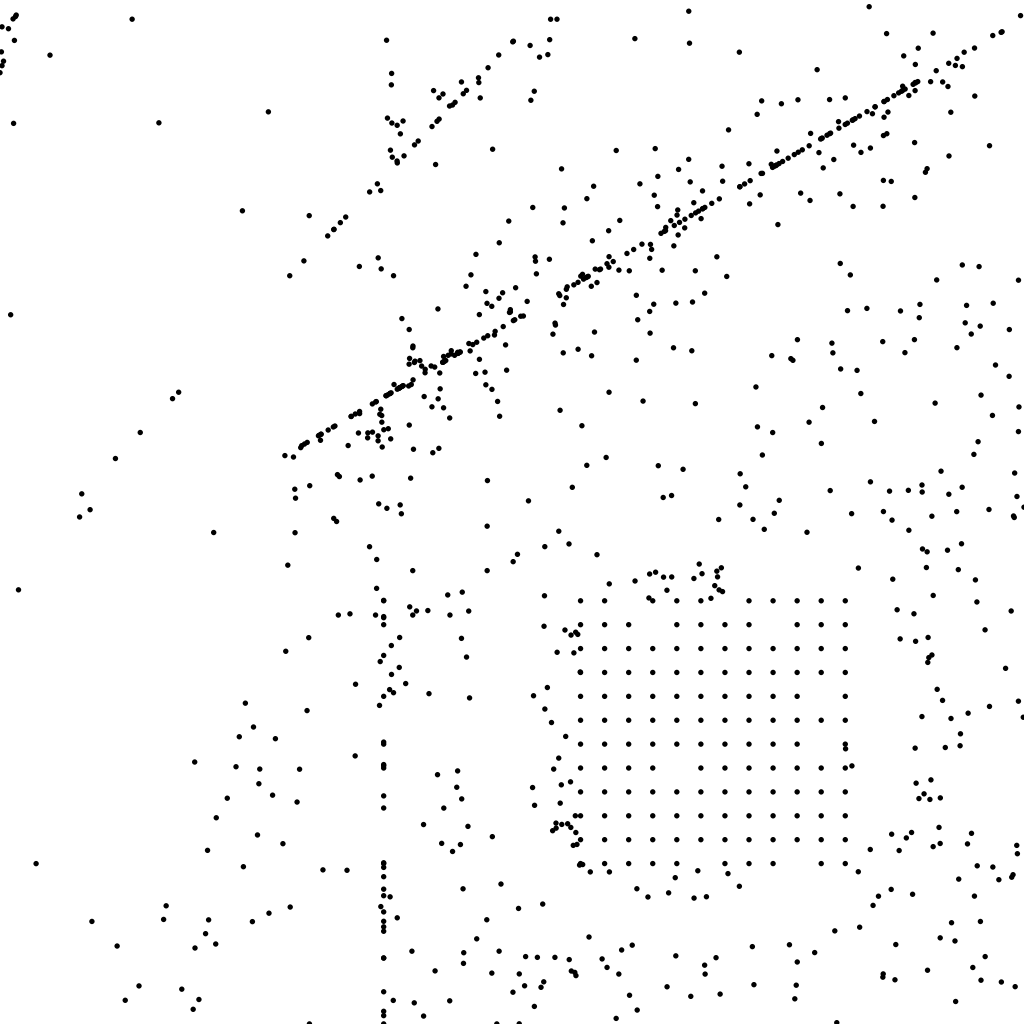}
    \hfill
    \includegraphics[width=0.32\textwidth,frame]{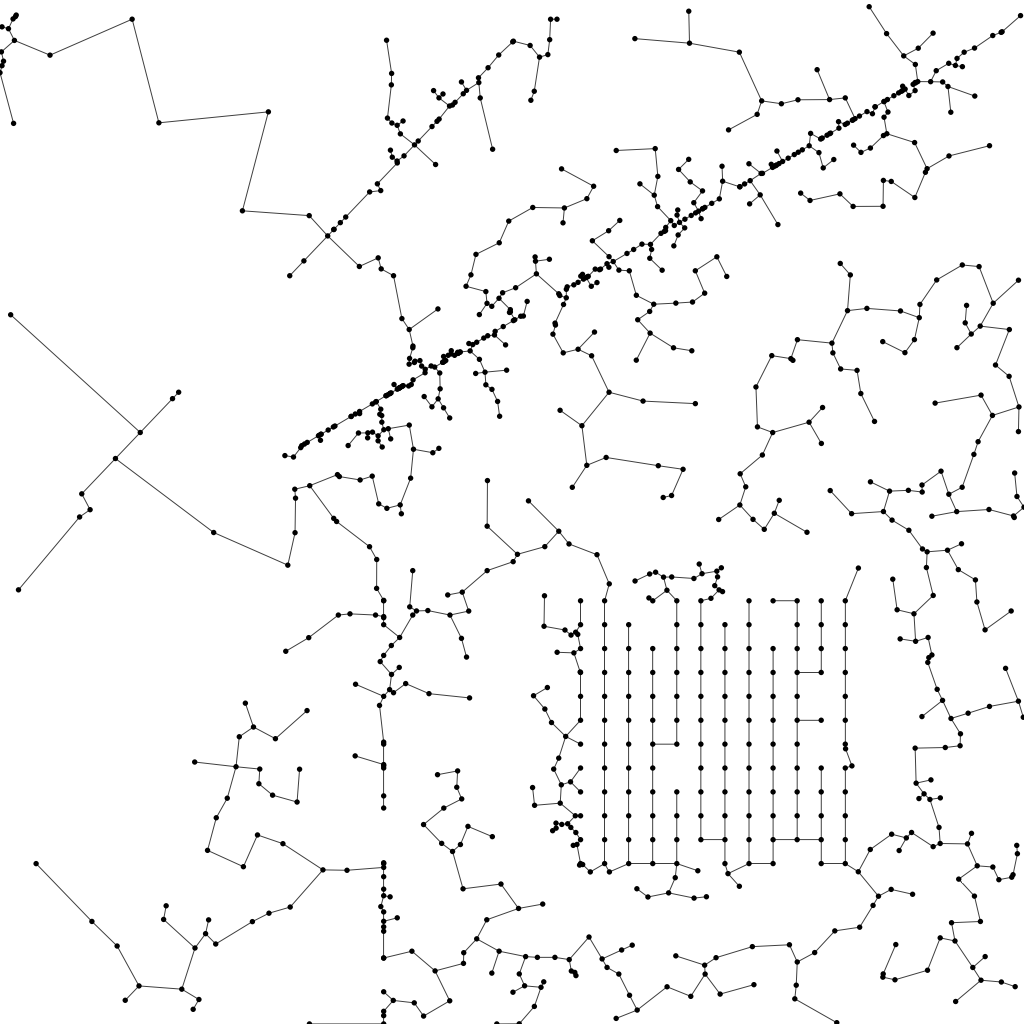}
    \hfill
    \includegraphics[width=0.32\textwidth,frame]{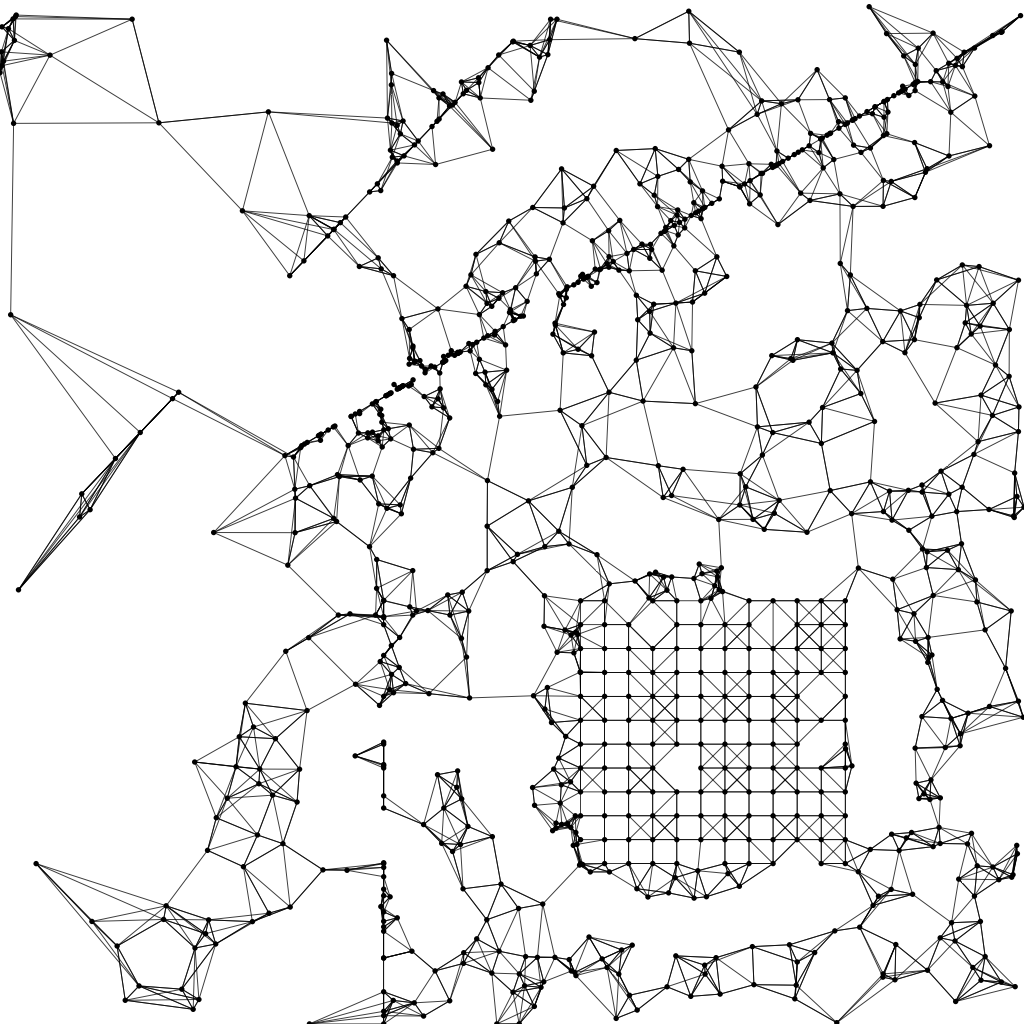}
    \caption{Exemplary visual representations of two TSP instances in terms of point cloud only (left), a minimum spanning tree (center) and the 5-nearest-neighbor graph (right).
    The top row shows the instance for which the highest mean PAR10 score was achieved by EAX, the bottom row shows the respective counterpart of LKH.
    }
    \vspace*{-0.35cm}
    \label{fig:tsp_image_data}
\end{figure}

The state-of-the-art TSP-related feature sets~\cite{Hutter201479,MBB2012LocalSearchTSP,Pihera14} consist of hundreds of hand-crafted features. Features range from statistics (mean, variance etc.) of edge lengths, angles of nearest neighbors, to more sophisticated features based on Minimum Spanning Trees (MST) or $k$-Nearest-Neighbor-Graphs ($k$-NNG).
Fig.~\ref{fig:tsp_image_data} depicts visual impressions of MSTs and $k$-NNGs on two evolved instances; these images will be a key ingredient to the neural network in Section~\ref{sec:as_dnn}.

Due to the size of feature sets, in the context of algorithm selection, automated feature selection methods have shown their suitability for automatically filtering a (small) subset of discriminating features \cite{KKBHTLeveragingTSP}. Regardless of the sophistication of the feature selection method at hand, feature selection needs to cope with an exponentially sized search space of the underlying subset-selection problem. Hence, in order to assist our feature-based model fitting we conduct a simple univariate exploratory data analysis in order to identify a lucid subset of adequate features \emph{a-priori}. To this end we adopt a simple greedy heuristic. First, all features $f$ are scaled to $[0,1]$ to allow for a fair comparison across the features. Then, we perform a two-sided non-parametric Wilcoxon-Mann-Whitney test~\cite{Mann1947WilcoxonTest} at significance level $\alpha=0.05$ per feature $f$, to check the null hypothesis that the distributions of the EAX instances and the LKH instances with respect to $f$ differ by a location shift equal to zero. We extract the 15 most relevant features according to the smallest $p$-values. This procedure was done once for the entire benchmark set of salesperson features as the most comprehensive feature set (see Sec. \ref{sec:as_classical}) and was repeated for subsets of \enquote{hardest} instances of decreasing size. Hardest in this context relates to (a) each 300 and 150 hardest instances for each solver with respect to mean PAR10-score, or (b) the ratio of mean PAR10-scores. Both methods pursue to reduce the benchmark set to instances with maximal performance differences.
\begin{figure}[ht]
    \centering
    \includegraphics[width=\textwidth]{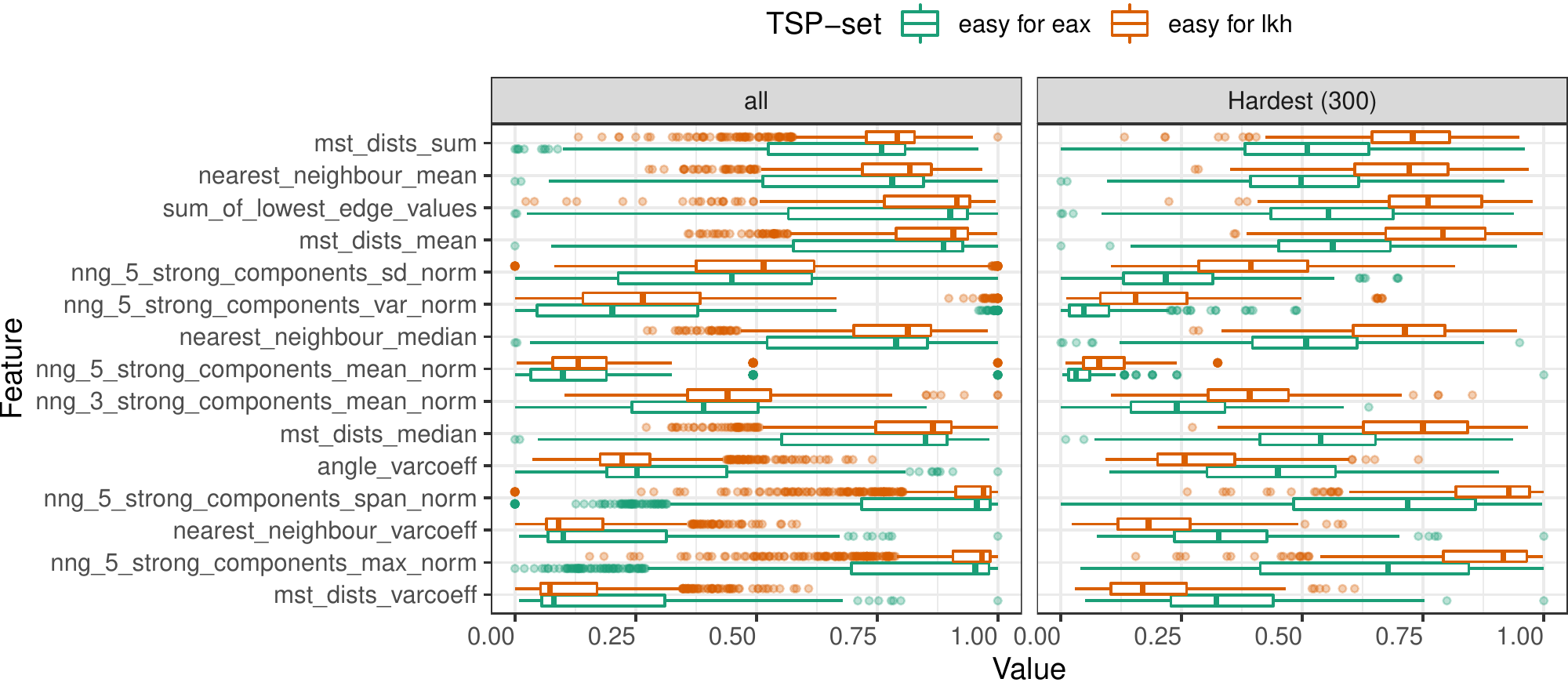}
    \caption{Distribution of 15 best features according to the significance-test based feature importance method on all instances (left) and the 300 hardest instances with respect to mean PAR10 performance (right).}
    \label{fig:boxplots_features}
\end{figure}

Fig.~\ref{fig:boxplots_features} shows the distribution of selected features both across the whole benchmark set and the subset of each $300$ most difficult instances with respect to mean PAR10-score for each solver (marked with crosses in Fig.~\ref{fig:scatterplot_performance}).
Noticeably, the 15 most relevant features are identical for both sets. They are mainly composed of summary statistics on strong connected components of the nearest neighbor graph (nng\_*) and properties based on minimum spanning trees (mst\_*). This is very much in line with crucial features identified in most TSP-related AS-studies~\cite{MBB2012LocalSearchTSP,Mersmann2013,Pihera14,Bossek2016EvolvingInstances,Bossek2016UnderstandingCharacteristics,KKBHTLeveragingTSP} by sophisticated variable importance measurement. These features seem plausible since both MSTs and NNGs capture the global structure, e.g., existence of clusters etc., very well. A close look at Fig.~\ref{fig:boxplots_features} suggests, that instances that are easy for EAX cover a wider range of values which is derived from wider (green) boxes for the features while easy instances for LKH show much more narrow (orange) boxes, with many strong outliers though. The right hand plot, however, shows that the hardest instances seem to be better separable with the features. For instance, for the features in the top 4 rows we observe that $75\%$ of feature values for LKH-friendly instances are higher than $75\%$ of the respective values for EAX-friendly instances.


\section{Classical Algorithm Selection}
\label{sec:as_classical}

Fig.~\ref{fig:scatterplot_performance} reveals very complementary performances of EAX and LKH, which gives us reason to assume that automated AS might work well in this setting. Further, as outlined in the univariate analysis of the TSP features (see Sec.~\ref{sec:feature_identification}), the features also indicate their potential for distinguishing instances that are beneficial for EAX from instances for which LKH is preferable. As previous works \cite{kotthoff2015,KKBHTLeveragingTSP} already confirmed the effectivity of feature-based AS, we adopted their experimental setup -- and only slightly modified it to the scenario at hand. Below, we will outline the considered machine learning algorithms, feature sets, as well as feature selection strategies, which have been used for training our final selectors.

\subsection{Experimental Setup}

All our candidate AS models are trained using PAR10 \cite{KBT2018Parameterization} as performance measure and assessed with a 10-fold cross-validation (CV). As indicated by Fig.~\ref{fig:scatterplot_performance}, it is much more difficult for LKH to perform well on the instances that were evolved in favor of EAX (green points), rather than vice versa (orange). Therefore, the selectors will likely have a bias towards EAX instances. To adjust for this bias, we additionally tune the classification threshold for all trained models. 

For training the potential automated AS models, we considered four different classifiers \cite{hastie2009} using the R-package \texttt{mlr} \cite{bischl2016mlr}: decision trees \cite{rpart2019}, random forests \cite{randomForest2002}, support vector machines \cite{kernlab2004} and gradient boosting \cite{xgboost2019}. Each of them is trained using three different feature sets: the UBC features from Hutter et al.~\cite{Hutter201479}, a large set of features by Pihera and Musliu \cite{Pihera14}, as well as the TSP features from the R-package \texttt{salesperson}\footnote{The R-package \texttt{salesperson} is an efficient and more comprehensive extension of the \texttt{tspmeta} feature generator by \cite{Mersmann2013}.}~\cite{salesperson2017}. 
The salesperson features provide the up-to-now most comprehensive collection of features; in fact, they are a strict superset of the Pihera and tspmeta features \cite{Mersmann2013}. On the other hand, the UBC and Pihera features led to the best performing algorithm selectors in previous works \cite{kotthoff2015,KKBHTLeveragingTSP} -- which did not consider the salesperson features as the package did not exist back then. Note that there is a large overlap across the three considered feature sets as outlined in \cite{KerschkeHNT2019AutomatedAlgorithm}.
To reduce the noise within and redundancy between the features, we additionally created five small subgroups from the \texttt{salesperson} feature set, consisting of 15 features each (see Sec.~\ref{sec:feature_identification} for details).

In addition to the 32 potential AS models described above (8 feature sets $\times$ 4 learners), the respective feature sets were further reduced using three automated feature selection strategies: sequential floating forward selection (sffs), sequential floating backward selection (sfbs), and -- for the reduced feature sets -- exhaustive search of the 15 features \cite{kohavi1997wrappers,KerschkeHNT2019AutomatedAlgorithm}. This resulted in 84 further candidate selectors.


\subsection{Findings}

Tab.~\ref{tab:overview_par10_classical_as} summarizes the averaged PAR10 performances of all 116 considered AS models, with the best achieved scores highlighted in red. Of course, all shown PAR10-scores already include the costs for the computation of the TSP features. On average those costs account for merely 0.7s at most.
According to the listed performances of the best models (61.21s), the SBS (67.47s) and the VBS (4.92s), the best classical AS approaches are able to reduce the SBS-VBS-gap by 10\%.

The best found selectors are random forests, which reduced the top 15 features that they were given initially, to the following four features: the sum, arithmetic mean, median and coefficient of variation of the distances of the MST. Moreover, the tuned thresholds varied from 4\% to 33\% across the ten folds, implying that EAX has always been selected once the model predicted EAX with a probability of at least 33\%. In consequence, out of all instances, in which LKH was actually the faster solver, it has only been selected 151 times (corresonding to roughly 30\%). On the other hand, LKH has only been (wrongly) picked in 4\% of the cases, in which EAX would have been the correct choice.

When starting with the full feature sets from Pihera, UBC and salesperson, a support vector machine based on a subset of the UBC features achieved the best performance (printed in bold in Tab.~\ref{tab:overview_par10_classical_as}). In fact, the PAR10-score of 56.67s is only slightly worse than the one of our best selector(s). Noticeably, the SVM also relied on MST features only: the arithmetic mean and standard deviation of the lengths of the edges in the MST, as well as the skewness of its node degrees.

\begin{table}[t!]
\caption{Overview of all PAR10 results using the classical feature-based AS approach. The best PAR10-scores are highlighted in red. Note that we did not perform exhaustive feature selection on the non-reduced feature sets due to enormous computational costs.}
\label{tab:overview_par10_classical_as}
\centering
\renewcommand{\tabcolsep}{9pt}
\renewcommand{\arraystretch}{1.1}
\begin{tabular}{cccccccccc}
\toprule
  \multirow{2}{*}{\makecell{\bfseries Feature Sel.\\[4pt]\bfseries Method}} & \multirow{2}{*}{\makecell{\bfseries ML\\[4pt]\bfseries Algor.}} & \multicolumn{3}{c}{\textbf{All Features}} & \multicolumn{5}{c}{\textbf{Top 15 Features}} \\
\cmidrule(l{2pt}r{2pt}){3-5}
\cmidrule(l{2pt}r{2pt}){6-10}
& & \bfseries Pihera & \bfseries \;UBC\; & \bfseries \;Sales.\; & \bfseries all & \bfseries r150 & \bfseries r300 & \bfseries s150 & \bfseries s300\\
  \midrule
  \multirow{4}{*}{none} & rpart & 67.25 & 69.35 & 67.70 & 67.32 & 67.32 & 67.52 & 68.12 & 67.32 \\
   & rf & 59.09 & 61.56 & 62.22 & 61.21 & 61.99 & 60.14 & 61.51 & 62.30 \\
   & xgboost & 65.40 & 67.00 & 67.57 & 65.53 & 65.53 & 64.51 & 65.78 & 65.53 \\
   & ksvm & 61.34 & 61.55 & 63.09 & 61.98 & 61.98 & 60.54 & 61.26 & 61.98 \\ \midrule
  \multirow{4}{*}{sffs} & rpart & 66.20 & 67.92 & 67.26 & 66.82 & 66.82 & 66.82 & 66.82 & 66.82 \\
   & rf & 325.27 & 58.42 & 62.77 & 60.70 & 59.74 & \textcolor{red}{\textbf{56.29}} & \textcolor{red}{\textbf{56.29}} & \textcolor{red}{\textbf{56.29}}\\
   & xgboost & 65.45 & 66.60 & 62.61 & 64.20 & 64.20 & 64.20 & 64.20 & 64.20 \\
   & ksvm & 66.42 & \textbf{56.67} & 60.57 & 62.38 & 62.38 & 62.22 & 62.06 & 62.38 \\ \midrule
  \multirow{4}{*}{sfbs} & rpart & 67.16 & 66.10 & 65.61 & 67.12 & 67.12 & 67.19 & 68.12 & 67.12 \\
   & rf & 60.01 & 61.56 & 63.24 & 62.05 & 61.95 & 59.62 & 62.19 & 62.17 \\
   & xgboost & 65.47 & 67.00 & 67.57 & 65.88 & 63.87 & 64.77 & 61.87 & 65.88 \\
   & ksvm & 61.25 & 63.07 & 62.88 & 60.66 & 60.62 & 59.40 & 62.75 & 60.47 \\ \midrule
  \multirow{4}{*}{exh.} & rpart &  &  &  & 67.19 & 67.19 & 66.82 & 67.10 & 67.19 \\
   & rf &  &  &  & 86.70 & 59.68 & \textcolor{red}{\textbf{56.29}} & 60.57 & 63.58 \\
   & xgboost &  &  &  & 61.75 & 64.28 & 63.02 & 64.09 & 63.02 \\
   & ksvm &  &  &  & 62.16 & 62.16 & 62.16 & 62.07 & 62.16 \\
  \bottomrule
  \multicolumn{10}{l}{
  \makecell{
  \textbf{Note}: r300 (s300) indicates the set of 300 hardest instances w.r.t.~PAR10-ratio (PAR10-score).
  } 
  } 
\end{tabular}
\end{table}

\begin{figure}[t!]
    \centering
    \includegraphics[width=0.49\textwidth]{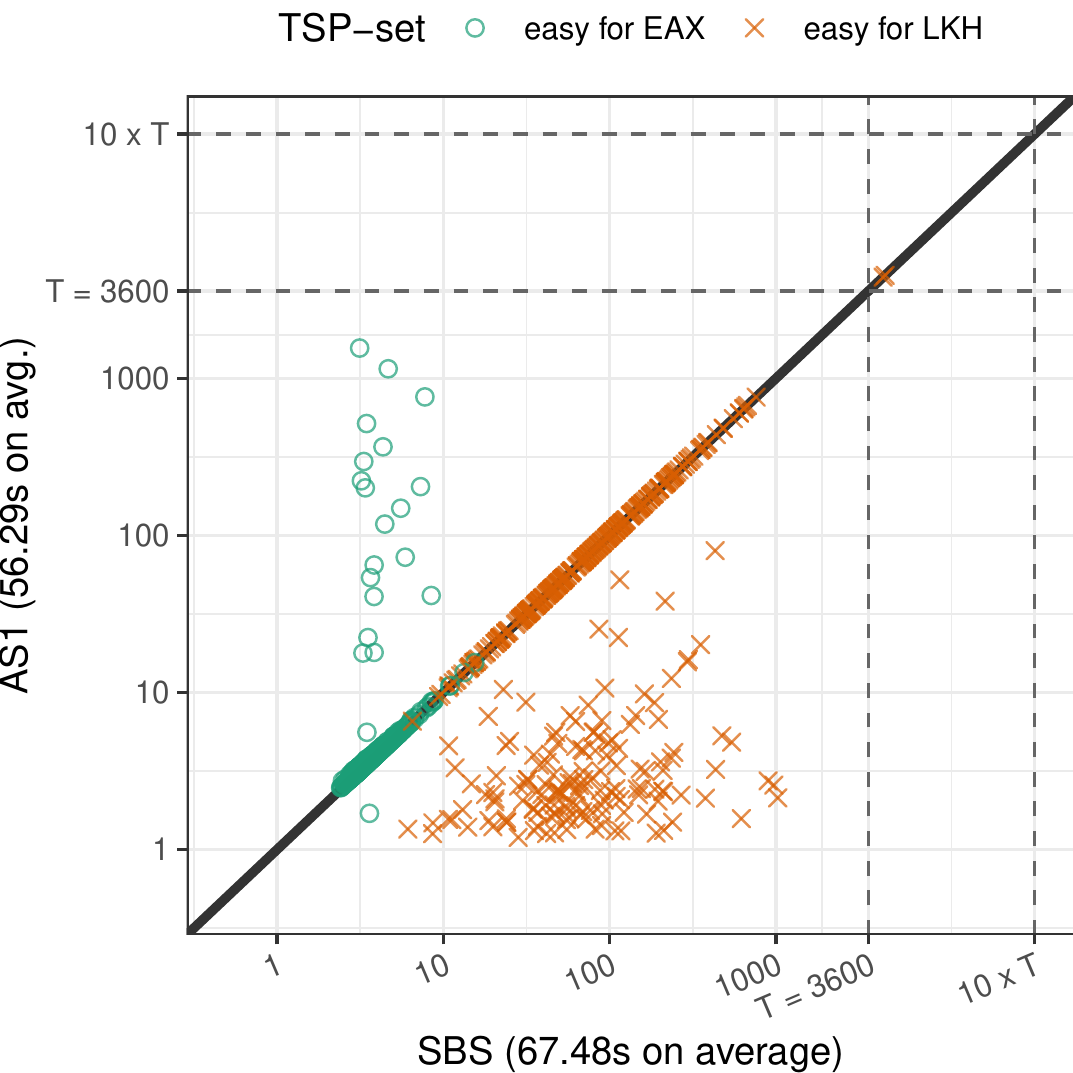}
    \hfill
    \includegraphics[width=0.49\textwidth]{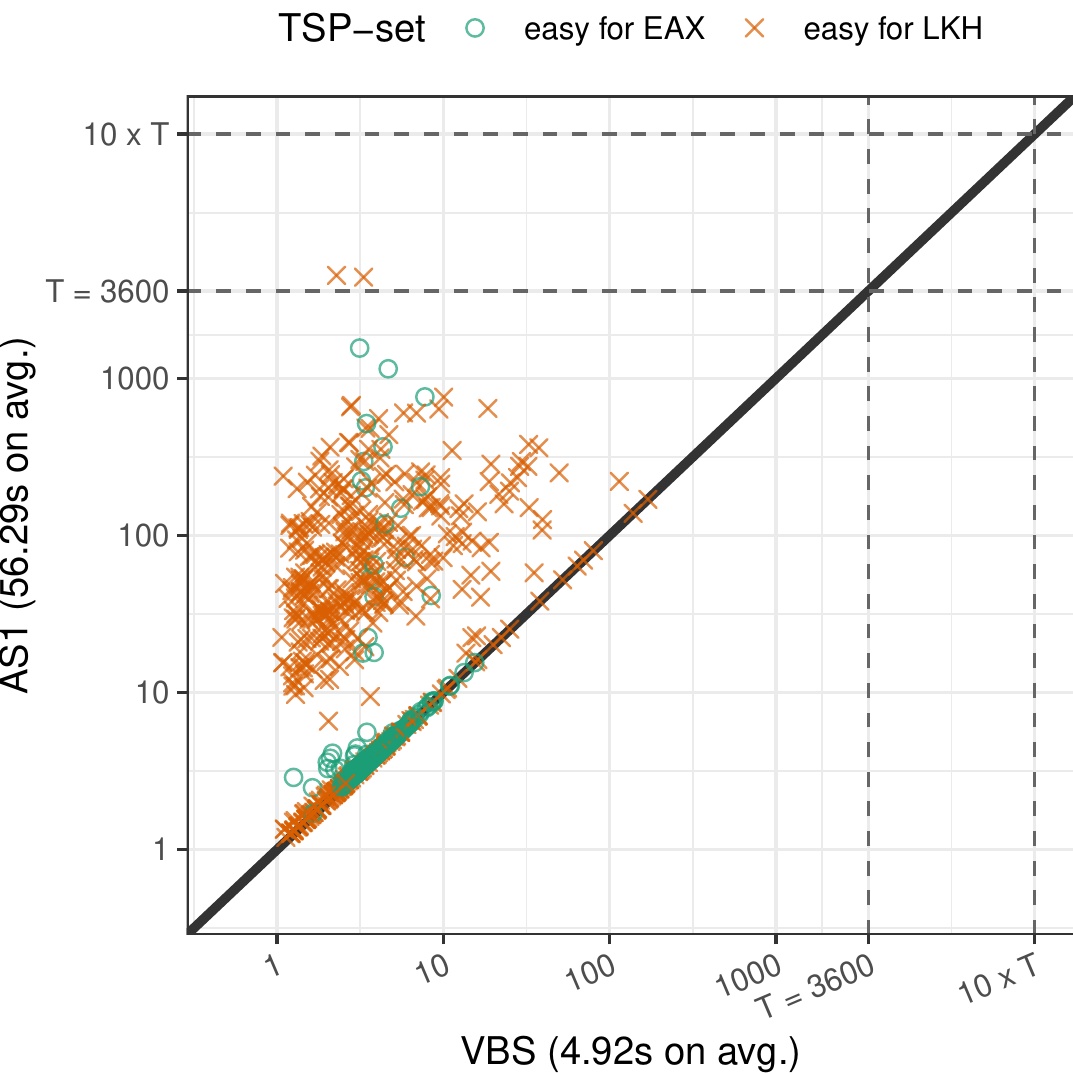}
    \caption{PAR10-scores (log-scaled) of the best classical AS model reveal the improvement of the best selector over the SBS (left), and the gap towards the VBS (right).}
    \label{fig:scatter_classic_as_par10}
\end{figure}

Our findings are also confirmed by the left image of Fig.~\ref{fig:scatter_classic_as_par10} as only few observations (20) are located above the diagonal. However, looking at the right image, it becomes nearly obvious that the selector is still quite far away from the performance of the VBS, as shown by the many misclassifications (381 out of all 1\,000 instances) above the diagonal.

In an attempt to better understand the reason for these rather small thresholds, we investigated the misclassification costs in detail. \emph{Prior} to tuning the threshold, LKH was predicted 154 times when EAX would have been the correct choice -- and each of those misclassifications caused (on average) an overhead of 4\,423.89s. In contrast, the 128 cases in which EAX was predicted instead of LKH only came with an average penalty of 95.95s. \emph{After} tuning the thresholds, each of the 20 wrong predictions of LKH caused only 290.04s -- compared to the average penalty of 126.07s for the 361 wrong predictions of EAX. Thus, by being rather conservative and only predicting LKH in cases, where the model is highly certain, the selector was able to reduce the misclassification costs significantly.

As our results indicate, the common feature-based approaches are able to improve over the SBS. However, it is also noticeable that the currently available features still have a very hard time in extracting sufficient information from the TSP instance to reliably predict the better solver. Therefore, we will test the suitability of deep learning neural networks as an alternative or supporting means for automated algorithm selection.


\section{Deep Learning Based Approach}
\label{sec:as_dnn}

As demonstrated in the previous section, feature-based AS methods can outperform the SBS. However, these models come with three major drawbacks: they (1) are hand-crafted in a tedious process, (2) partly require time-consuming calculations, and (3) are problem tailored (see Sec.~\ref{sec:introduction}).
To overcome these issues, we propose a novel, and sophisticated feature-free approach that is based on so-called \textit{Convolutional Neural Networks} (CNN) \cite{lecun1995convolutional}.

\subsection{Experimental Setup}
To train CNN based AS models that are independent of the commonly used TSP features, we will produce different visual representations of our TSP instances (see Fig.~\ref{fig:tsp_image_data}) and use them for training the deep learning networks.
Those images are created with a resolution of $512 \times 512$ pixels and the coordinates of the instances are scaled to fill out the entire image as exemplarily shown in the two images in the left column of Fig.~\ref{fig:tsp_image_data}.
In addition to these point clouds, images of corresponding \textit{Minimum Spanning Trees} (MST, second column of Fig.~\ref{fig:tsp_image_data}) and \textit{$k$-Nearest-Neighbor Graphs} ($k$-NNG, right column of Fig.~\ref{fig:tsp_image_data}) with $k = 5$ were generated. We chose MST and $5$-NNG as additional visual representations because we found in Section~\ref{sec:feature_identification} that the 15 most important features are based almost exclusively on MST and $5$-NNG graphs. In the $5$-NNGs, not only mutual (strong) connections but also one-sided (weak) links were considered, in which one city belongs to the nearest neighbor set of another city, but not vice versa.


Admittedly, only networks whose generation was based exclusively on point clouds can be described as \emph{feature-free}. For a better comparison, however, we have additionally evaluated \emph{feature-based} networks that were trained with images of the corresponding MST and $5$-NNG. Hence, we considered two different scenarios for our network-based approaches. In the first scenario (S1), the networks were trained based on (a) point clouds of the cities (Points), (b) MST images, and (c) $5$-NNG images. In scenario (S2), we combined (a) the scatterplots with the MST images (as two input channels), and (b) the scatterplots with the MST and $5$-NNG images (as three input channels). As the costs for generating the images are insignificant, we have not taken their generation time into account when computing the PAR10 scores of the deep learning models. For larger instances, though, these times would have to be taken into account.

\begin{figure}[t!]
    \centering
    \begin{tikzpicture}[scale=1.0]
        \node[draw=none,rotate=90, minimum width = 2.5cm] (InputV) at (0,0) {Visual Rep.};

        \node[draw,rotate=90, minimum width = 1.8cm] (CNN1) at (1,0) {3x3 32};
        \node[draw,rotate=90, minimum width = 1.8cm] (CNN2) at (2,0) {3x3 32};
        \node[draw,rotate=90, minimum width = 1.8cm] (CNN3) at (3,0) {3x3 64};
        \node[draw,rotate=90, minimum width = 1.8cm] (CNN4) at (4,0) {3x3 64};
        \node[draw,rotate=90, minimum width = 1.8cm] (CNN5) at (5,0) {3x3 128};
        \node[draw,rotate=90, minimum width = 1.8cm] (CNN6) at (6,0) {3x3 128};
        \node[draw,rotate=90, minimum width = 1.8cm] (CNN7) at (7,0) {3x3 256};
        \node[draw,rotate=90, minimum width = 1.8cm] (CNN8) at (8,0) {3x3 256};
        \node[draw,rotate=90, minimum width = 2.5cm] (GPL)  at (9,0) {Global Pool.};
        \node[draw,rotate=90, minimum width = 2.5cm] (DP)  at (10,0) {Dropout 25\%};
        \node[draw,rotate=90, minimum width = 2.5cm] (L2)  at (11,0) {Linear 2};

        \draw[->, line width = 0.05cm] (InputV) to (CNN1);
        \draw[-, line width = 0.02cm] (CNN1) to (CNN2);
        \draw[-, line width = 0.02cm] (CNN2) to (CNN3);
        \draw[-, line width = 0.02cm] (CNN3) to (CNN4);
        \draw[-, line width = 0.02cm] (CNN4) to (CNN5);
        \draw[-, line width = 0.02cm] (CNN5) to (CNN6);
        \draw[-, line width = 0.02cm] (CNN6) to (CNN7);
        \draw[-, line width = 0.02cm] (CNN7) to (CNN8);
        \draw[-, line width = 0.02cm] (CNN8) to (GPL);
        \draw[-, line width = 0.02cm] (GPL) to (DP);
        \draw[-, line width = 0.02cm] (DP) to (L2);

        \node[draw=none] ()  at (-0.3,-1.5) {Dilation};
        \node[draw=none] ()  at (1,-1.5) {1};
        \node[draw=none] ()  at (2,-1.5) {1};
        \node[draw=none] ()  at (3,-1.5) {2};
        \node[draw=none] ()  at (4,-1.5) {2};
        \node[draw=none] ()  at (5,-1.5) {3};
        \node[draw=none] ()  at (6,-1.5) {3};
        \node[draw=none] ()  at (7,-1.5) {1};
        \node[draw=none] ()  at (8,-1.5) {1};

        \node[draw=none] ()  at (-0.3,-2) {Strides};
        \node[draw=none] ()  at (1,-2) {1};
        \node[draw=none] ()  at (2,-2) {2};
        \node[draw=none] ()  at (3,-2) {1};
        \node[draw=none] ()  at (4,-2) {2};
        \node[draw=none] ()  at (5,-2) {1};
        \node[draw=none] ()  at (6,-2) {2};
        \node[draw=none] ()  at (7,-2) {1};
        \node[draw=none] ()  at (8,-2) {1};
    \end{tikzpicture}
    \vspace*{-0.2cm}
    \caption{The chosen neural architecture. All convolutional blocks include a \textit{Group Normalization Layer} and a \textit{Rectified Linear Unit} activation (Conv $\rightarrow$ GN $\rightarrow$ ReLU). The strides are used to reduce the feature maps' dimensions and the dilation are used to increase the receptive fields without adding additional parameters.}
    \label{fig:cnn_topology}
\medskip
\includegraphics[width=0.25\textwidth]{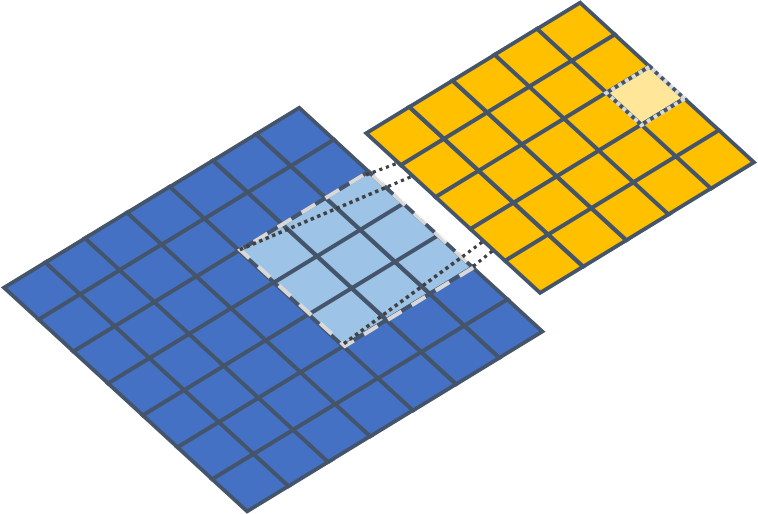}
\hspace{0.5cm}
\includegraphics[width=0.25\textwidth]{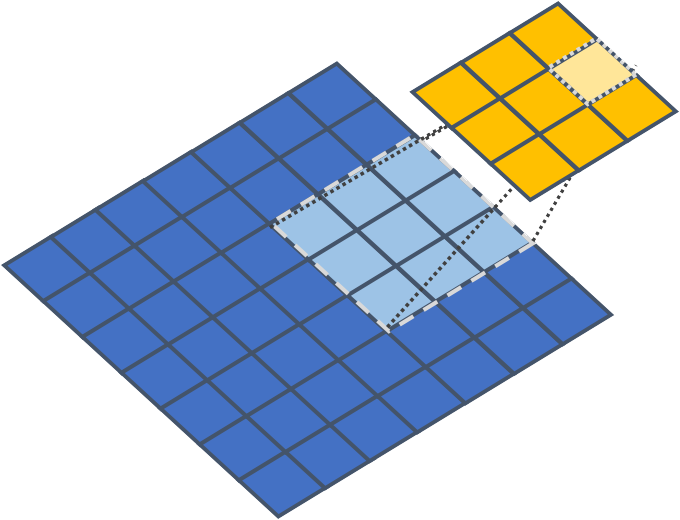}
\hspace{0.5cm}
\includegraphics[width=0.25\textwidth]{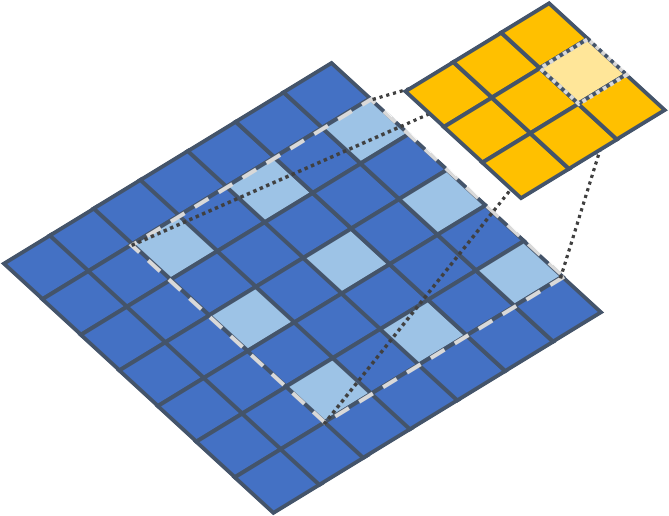}
\caption{Exemplary visualization of the operation principles of \textit{Convolutional Neural Networks}. Left: Normal CNN layer with $Dilation=1$ and $Strides=1$. Middle: CNN layer with $Strides=2$ (reducing the output size by half). Right: CNN Layer with $Dilation=2$ (increasing the receptive field without adding additional weights).}
\label{fig:cnn_working}
\end{figure}

To process the visual representations of the instances, we used eight stacked convolutional layers (see Fig.~\ref{fig:cnn_topology}). Three of them used $Strides=2$ to reduce the size of the feature maps and four of them used $Dilation=\{2,3\}$-Kernels to enlarge the receptive fields and, thus, gain a larger view of the instances (see Fig.~\ref{fig:cnn_working} to compare the effects of Strides and Dilation). We used \emph{Rectified Linear Unit} (ReLU) \cite{GlorotBB11} as activation function for all layers except for the last linear layer, for which we used a \emph{Softmax} activation. To improve the training speed, the outputs of all convolutional layers are normalized by using \emph{Group Normalization} (GN) \cite{WuH20} with $G=8$. The GN layers are in-between the convolutional layers and the ReLU activation. For transition from the three-dimensional convolutional ($Width \times Height \times Channels$) layers to the one-dimensional linear layer, a \emph{Global Average Pooling Layer} (GPL) \cite{lin2013network} is used. After the GPL, a Dropout (DP) \cite{srivastava2014dropout} layer with 25\% dropout is added to improve regularization. The final layer is a single, linear layer with two output neurons -- one for EAX and one for LKH (see Fig. \ref{fig:cnn_topology}). Last, we used 10-fold cross-validation to evaluate the performance of the neural networks. The folds were the same as for the classical feature-based approach. All networks were trained using mini-batches of eight, Adam \cite{KingmaB14} as optimizer and \textit{Cross-Entropy} \cite{MannorPR05} as loss function. Note that neural networks are most commonly trained using \emph{Stochastic Gradient Descent} \cite{robbins1951stochastic}, which strongly differs from the training methods used in Section~\ref{sec:as_classical}.

\subsection{Findings}
The best performing classical approach achieved a mean PAR10-score of $56.29$ (see Tab.~\ref{tab:overview_par10_classical_as}). Our feature-free networks, which were trained exclusively using the points, achieved a mean PAR10-score of $56.31$ after tuning the threshold and thus a similar performance (see Tab.~\ref{tab:cnn_results_overview}) as the feature-based, classical approaches.


As stated before, we additionally investigated whether adding additional features to the networks could improve the models' performances. Therefore, we also trained feature-based networks using MST and 5-NNG images. As shown in Tab.~\ref{tab:cnn_results_overview}, both variants perform noticeably better.
Besides, we found that while the thresholds between the points and the MST models are rather similar, the thresholds of the NNG models are on average $10\%$ higher. Next, the thresholds of the Points and MST models range from $13\%$ to $34\%$, and $16\%$ to $35\%$, respectively, while the thresholds of the NNG models range from $3\%$ to $73\%$.
Thus, networks trained on the NNG images appear to be less stable.


Moreover, the networks based on the Points correctly predicted EAX in 91.8\% (and thus 459 times) of the cases, in which EAX was the better solver, compared to only 22.6\% (113 cases) for the LKH-friendly cases. This behavior likely results from the fact that a misclassification of an instance, which is favorable for LKH, is cheaper than a misclassification of an instance that is easier for EAX.
In contrast, the MST-based networks predict EAX in 91.6\% (458) and LKH in 27.2\% (136) cases, correctly.
Thus, compared to the feature-free networks, which were exclusively based on Points, the MST networks benefit from correctly identifying LKH-friendly instances -- without losing accuracy on the EAX-friendly instances.
Noticeably, in case of the 5-NNG networks, only 84.6\% (423) of the EAX-easy instances are classified correctly, compared to 34.8\% (174) among the LKH-easy instances. Thus, despite the improvements among the instances that are favorable for LKH, the PAR10 score of the NNG-based networks is inferior to the MST-based selector, as misclassifying EAX-easy instances is more expensive.


\begin{table}[t!]
\caption{PAR10-scores of our deep neural networks for the different scenarios and across all ten CV-folds. In addition, we list the values of the tuned thresholds (TH).}
\begin{center}
\renewcommand{\tabcolsep}{9pt}
\renewcommand{\arraystretch}{1.1}
\begin{tabular}{l cccccc cccc}
    \toprule
     & \multicolumn{6}{c}{\textbf{Scenario S1}}& \multicolumn{4}{c}{\textbf{Scenario S2}}\\
     \cmidrule(l{2pt}r{2pt}){2-7}
     \cmidrule(l{2pt}r{2pt}){8-11}
     & \multicolumn{2}{c}{Points} & \multicolumn{2}{c}{MST} & \multicolumn{2}{c}{NNG} & \multicolumn{2}{c}{Points + MST} & \multicolumn{2}{c}{Points+MST+NNG} \\
     \cmidrule(l{2pt}r{2pt}){2-3}
     \cmidrule(l{2pt}r{2pt}){4-5}
     \cmidrule(l{2pt}r{2pt}){6-7}
     \cmidrule(l{2pt}r{2pt}){8-9}
     \cmidrule(l{2pt}r{2pt}){10-11}
Fold & PAR10 & TH & PAR10 & TH & PAR10 & TH& PAR10 & TH & PAR10 & TH \\
    \midrule
1 & 34.40 & 0.19 & 30.91 & 0.27 & 31.35 & 0.64 & \textbf{27.13} & 0.21 & 32.95 & 0.14 \\
2 & 78.88 & 0.15 & 57.95 & 0.17  & 76.99 & 0.18  & \textbf{54.65} & 0.14 & 58.06 & 0.22  \\
3 & 67.43 & 0.14 & 64.14 & 0.19  & \textbf{57.94}  & 0.32  & 62.40 & 0.25 & 62.15 & 0.15 \\
4 & 50.50 & 0.13 & \textbf{45.89} & 0.19  & 59.35 & 0.03  & 54.92 & 0.19 & 56.58 & 0.13 \\
5 & \textbf{52.41} & 0.29 & 57.76 & 0.16  & 59.40 & 0.43  & 58.50 & 0.24 & 61.43 & 0.25 \\
6 & 58.07 & 0.15 & \textbf{45.28}  & 0.21  & 56.09 & 0.35  & 64.59 & 0.17 & 54.95 & 0.19  \\
7 & 50.32 & 0.40 & 58.91 & 0.29  & \textbf{42.49}  & 0.05  & 49.47 & 0.25 & 54.76 & 0.22  \\
8 & 36.89 & 0.23 & 35.35 & 0.20   & \textbf{26.08}  & 0.73  & 63.47 & 0.29 & 32.07 & 0.27 \\
9 & \textbf{93.07} & 0.31 & 95.44 & 0.23  & 95.41 & 0.41  & 96.89 & 0.16 & 93.57 & 0.27  \\
10 & \textbf{41.10} & 0.34 & 48.79 & 0.35  & 44.52 & 0.13  & 52.09 & 0.42 & 48.66 & 0.14 \\
\midrule
$\varnothing$& 56.31 & 0.23 & \textcolor{red}{\textbf{54.04}} & 0.23 & 54.96 & 0.33 & 58.41 & 0.23 & 55.52 & 0.20 \\
 \bottomrule
\end{tabular}
\end{center}
\label{tab:cnn_results_overview}
\vspace*{-0.5cm}
\end{table}

\begin{figure}[t]
    \centering
    \includegraphics[width=0.49\textwidth]{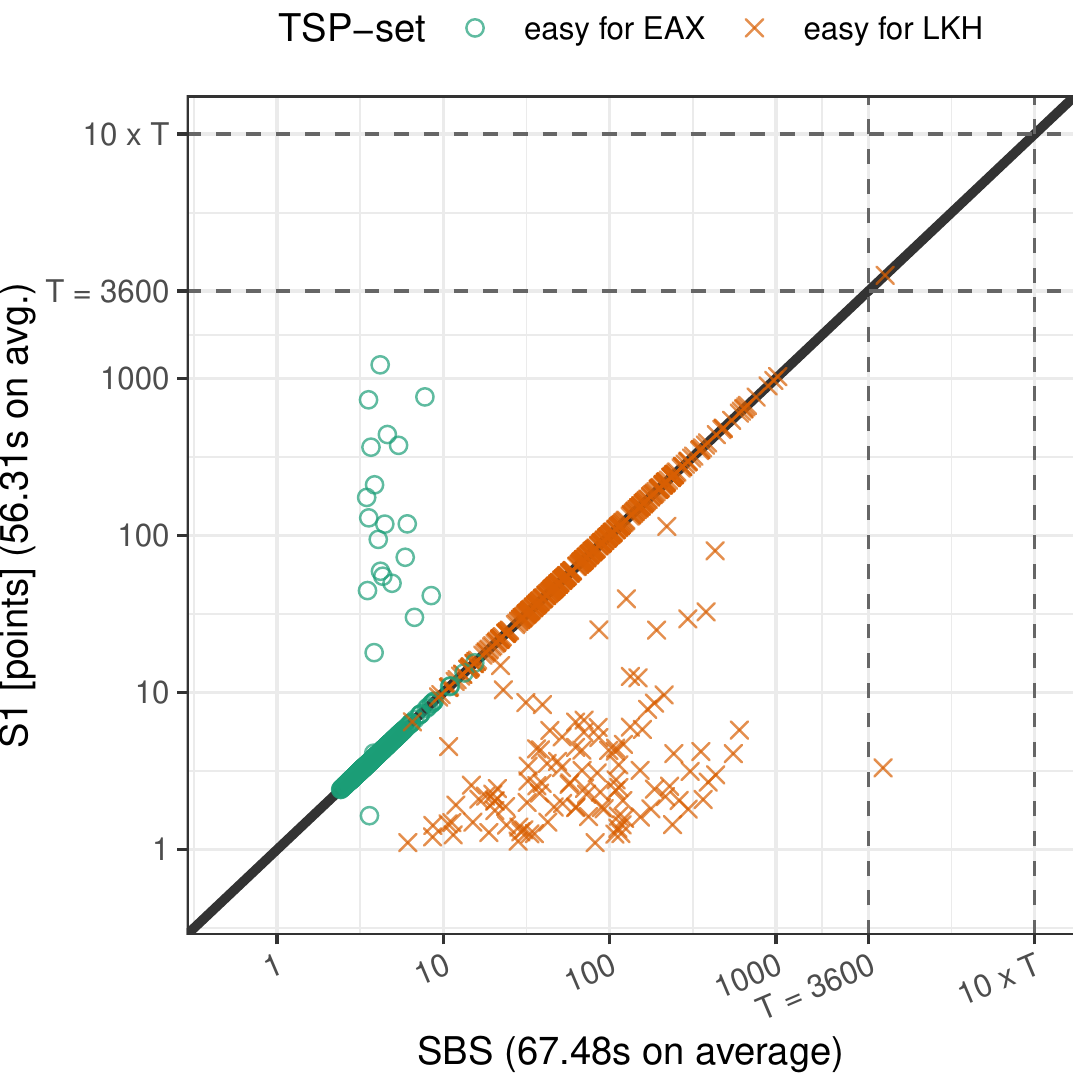}
    \hfill
    \includegraphics[width=0.49\textwidth]{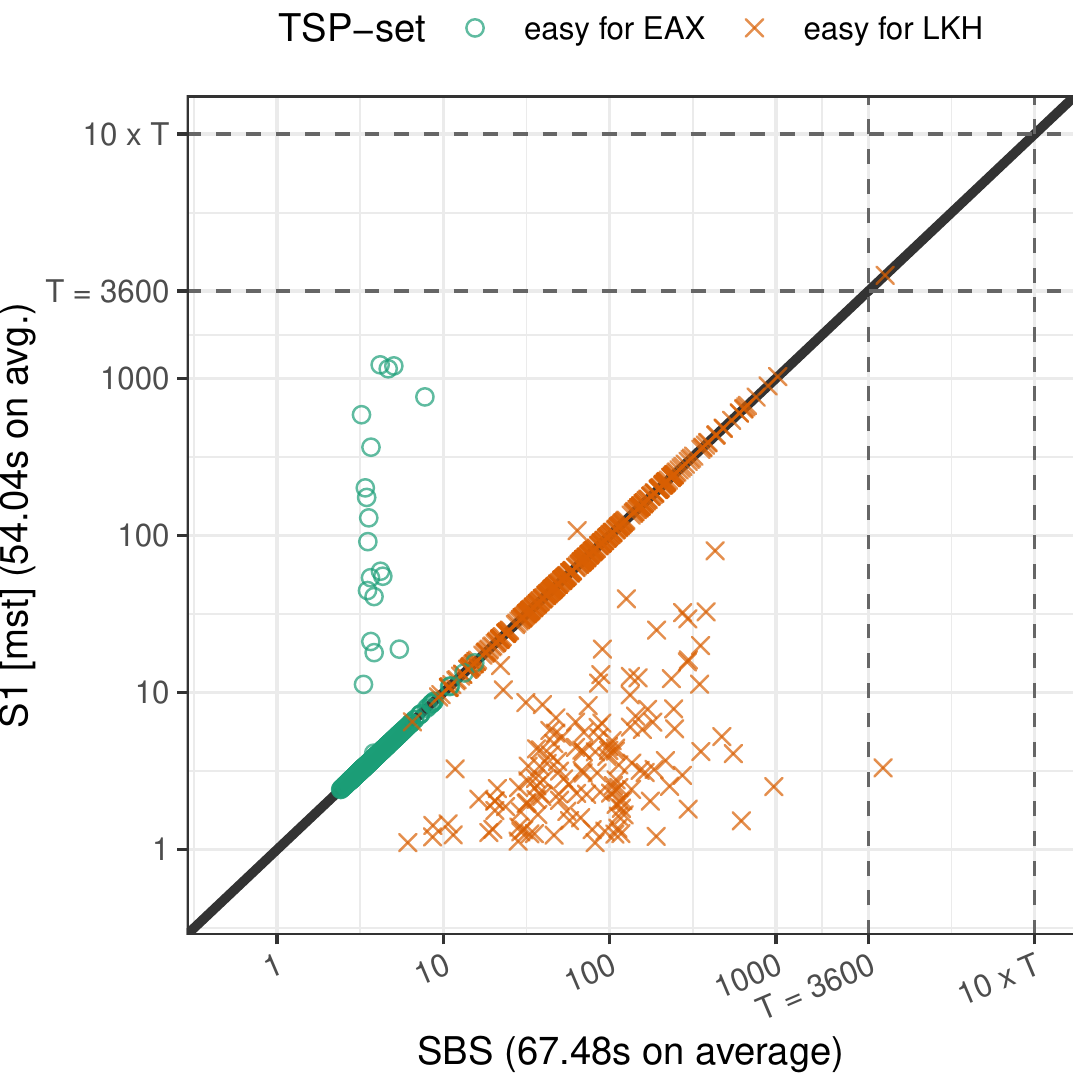}
    \vspace*{-0.1cm}
    \caption{PAR10-scores (log-scaled) of the points networks of scenario S1 (on the left) and the MST networks of scenario S1 versus the Single-Best-Solver (EAX).}
    \label{fig:scatter_dnn_as_par10}
    \vspace{-0.4cm}
\end{figure}

To investigate whether the combination of the three different input variants would lead to networks that achieve better performances when predicting EAX- and LKH-easy instances, we combined Points and MST, as well as Points, MST and NNG into two and three input channels (Scenario 2), respectively.
However, as shown in Tab.~\ref{tab:cnn_results_overview}, combining the visual representations does not improve the networks' overall performances. Also, while the network based on Points + MST classifies 31\% (155) of the LKH-easy instances correctly, the selector based on Points + MST + NNG only succeeds in 17\% (85) of the respective cases.
We further observed that the threshold values are quite similar to the ranges of the Points and MST models from scenario S1.

As visualized in Fig.~\ref{fig:scatter_dnn_as_par10}, the Points (left) and MST networks (right) prefer EAX over LKH -- as there are far more observations below the diagonal. This is also confirmed by the low thresholds (see Tab.~\ref{tab:cnn_results_overview}). Interestingly, the networks solely based on visual representations of the instance, even perform slightly better than the classical feature-based AS models -- but are still clearly inferior to the VBS. 



\section{Conclusions and Outlook}
\label{sec:conclusion}



The conducted experiments shed light on still existing shortcomings of classical feature-based per-instance algorithm selection on the TSP. While previous studies clearly reported successful approaches, the informative character of existing TSP feature sets reaches its limits for instances specifically evolved for maximum performance difference of the two state-of-the art heuristic solvers. Sophisticated mutation operators here lead to so far unobserved topological structures.
Despite outperforming the SBS, the gap to the performance of the oracle-like VBS cannot be closed substantially, even after utilization of sophisticated preprocessing and feature selection approaches.

However, it again becomes obvious that the minimum spanning tree and nearest neighbor structures of the points are most informative in discriminating solver performances. We build on this information and enrich a deep neural network approach based on images of the instance topology by specific images visualizing the minimum spanning tree and the nearest neighbor graph.
Most interestingly, our feature-free deep neural network nicely matches the performance of the quite complex classical AS approach (see Tab.~\ref{tab:wrapup}), despite being solely based on an image of the instance's points.

This proof-of-concept study thus shows the huge potential of deep learning, feature-free approaches in this domain which we will exploit in future studies by more sophisticated networks and altered loss functions specifically adapted to the designated performance indicators. Moreover, additional image channels will be added, e.g., in terms of heatmaps. Specific investigations have to be conducted with regard to the scaling behaviour of the approach, as image resolutions most probably will have to be carefully adapted to increasing instance sizes.

On the other hand, the observed limitations of classical TSP features show the necessity of enriching the library of TSP features by alternative sets which capture other kinds of — obviously important — instance structures. We will apply our approach to classical feature sets such as RUE or TSPLib as well for a comparison. However, it is specifically noteworthy that, in principle, the deep learning approach nicely generalizes to other graph-based optimization problems while instance features are almost exclusively tailored to the focused domain.

\begin{table}[t!]
\caption{Comparison of baseline (VBS and SBS) with our best models.
}
\label{tab:wrapup}
\centering
\renewcommand{\tabcolsep}{9pt}
\renewcommand{\arraystretch}{1.1}
\begin{tabular}{lccccccc}
\toprule
\multicolumn{1}{c}{\textbf{Measure}} & \multicolumn{2}{c}{\textbf{Baseline}} & \multicolumn{2}{c}{\textbf{Classical AS}} & \multicolumn{3}{c}{\textbf{Deep Learning AS}} \\
\cmidrule(l{2pt}r{2pt}){2-3}
\cmidrule(l{2pt}r{2pt}){4-5}
\cmidrule(l{2pt}r{2pt}){6-8}
& VBS & SBS~(EAX) & RF & SVM & Points & MST & Points~+~MST \\
\midrule
PAR10 & 4.92 & 67.47 & 56.29 & 56.67 & 56.31 & 54.04 & 55.52 \\
Accuracy & 1.00 & \phantom{0}0.49 & \phantom{0}0.62 & \phantom{0}0.60 & \phantom{0}0.57 & \phantom{0}0.59 & \phantom{0}0.61\\
F1-Score & 1.00 & \phantom{0}0.65 & \phantom{0}0.71 & \phantom{0}0.70 & \phantom{0}0.68 & \phantom{0}0.69 & \phantom{0}0.70 \\
\bottomrule
\end{tabular}
\end{table}

\subsection*{Acknowledgments}
The authors acknowledge support by the \href{https://www.ercis.org}{\textit{European Research Center for Information Systems (ERCIS)}}.

\bibliographystyle{arxiv}
\bibliography{arxiv}

\end{document}